\documentclass[sn-mathphys-ay]{sn-jnl}% Math and Physical Sciences Author Year Reference Style
\usepackage{graphicx}%
\usepackage{multirow}%
\usepackage{amsmath,amssymb,amsfonts}%
\usepackage{amsthm}%
\usepackage{mathrsfs}%
\usepackage[title]{appendix}%
\usepackage{xcolor}%
\usepackage{textcomp}%
\usepackage{manyfoot}%
\usepackage{booktabs}%
\usepackage{algorithm}%
\usepackage{algorithmicx}%
\usepackage{algpseudocode}%
\usepackage{listings}%
\usepackage{pdflscape}
\usepackage{lineno,hyperref}
\usepackage{tabularx}
\usepackage{makecell}
\usepackage{hhline}
\usepackage{makecell}
\usepackage{tabularx}
\usepackage{hhline}
\usepackage{colortbl}
\usepackage{adjustbox}
\usepackage{rotating}
\usepackage{longtable}
\usepackage{subcaption}
\usepackage{booktabs}
\usepackage{tikz}
\usepackage{pifont}

\usepackage{orcidlink}
\begin{document}

%\title[PEM-Color for GCP]{\vspace{-2.0cm}An island-parallel ensemble metaheuristic algorithm for large graph coloring problems}
\title[PEM-Color for GCP]{An island-parallel ensemble metaheuristic algorithm for large graph coloring problems}
%\author[inst1]{Tansel Dokeroglu}
%\address[inst1]{TED University, Software Engineering Department, Ankara, Turkey}
%\ead{tansel.dokeroglu@tedu.edu.tr}

%\author[inst2]{Tayfun Kucukyilmaz\corref{cor1}}
%\address[inst2]{Erasmus University, Rotterdam School of Management, Rotterdam, Netherlands}
%\ead{kucukyilmaz@rsm.nl}
%\cortext[cor1]{I am corresponding author}

\author[1]{\fnm{Tansel} \sur{Dokeroglu} \orcidlink{0000-0003-1665-5928}} \email{tansel.dokeroglu@tedu.edu.tr} 

\author*[2]{\fnm{Tayfun} \sur{Kucukyilmaz} \orcidlink{0000-0002-2551-4740}}\email{kucukyilmaz@rsm.nl}

\author[3]{\fnm{Ahmet} \sur{Cosar}
\orcidlink{0000-0002-3090-2254}} \email{cosar@metu.edu.tr}

%\author{Anonymous Author(s)}

%\affil*[1]{\orgdiv{Software Engineering Department}, \orgname{TED University},\orgaddress{\street{Ziya Gökalp Cd.}, \city{Çankaya}, \postcode{06420}, \state{Ankara}, \country{Turkey}}}

%\affil[2]{\orgdiv{Rotterdam School of Management}, \orgname{Erasmus University}, \orgaddress{\street{Burgemeester Oudlaan 50}, \city{Rotterdam}, \postcode{3062 PA}, \state{Rotterdam}, \country{Netherlands}}}

\abstract{Graph Coloring Problem (GCP) is an NP-Hard vertex labeling problem in graphs such that no two adjacent vertices can have the same color. Large instances of GCP cannot be solved in reasonable execution times by exact algorithms. Therefore, soft computing approaches, such as metaheuristics, have proven to be very efficient for solving large instances of GCP. In this study, we propose a new island-parallel ensemble metaheuristic algorithm (PEM-Color) to solve large GCP instances. Ensemble learning is a new machine learning approach based on combining the output of multiple models instead of using a single one. We use Message Passing Interface (MPI) parallel computation libraries to combine recent state-of-the-art metaheuristics: Harris Hawk Optimization (HHO), Artificial Bee Colony (ABC), and Teaching Learning Based (TLBO) to improve the quality of their solutions further. To the best of our knowledge, this is the first study that combines metaheuristics and applies to the GCP using an ensemble approach. We conducted experiments on large graph instances from the well-known DIMACS benchmark using 64 processors and achieved significant improvements in execution times. The experiments also indicate an almost linear speed-up with a strong scalability potential. The solution quality of the instances is promising, as our algorithm outperforms 13 state-of-the-art algorithms.}




\maketitle

%\textbf{CrediT Contributor Roles:} 

%\textbf{Conceptualization:} Tansel Dokeroglu; \textbf{Data Curation:} Tansel Dokeroglu; \textbf{Formal Analysis:} Tansel Dokeroglu, Tayfun Kucukyilmaz; \textbf{Methodology:} Tansel Dokeroglu, Tayfun Kucukyilmaz; \textbf{Software:} Tansel Dokeroglu; \textbf{Visualisation:} Tansel Dokeroglu, Tayfun Kucukyilmaz; \textbf{Writing:} Tansel Dokeroglu, Tayfun Kucukyilmaz

\section{Introduction}

\begin{figure}[t]
\begin{center}  
\includegraphics[width=0.5\textwidth]{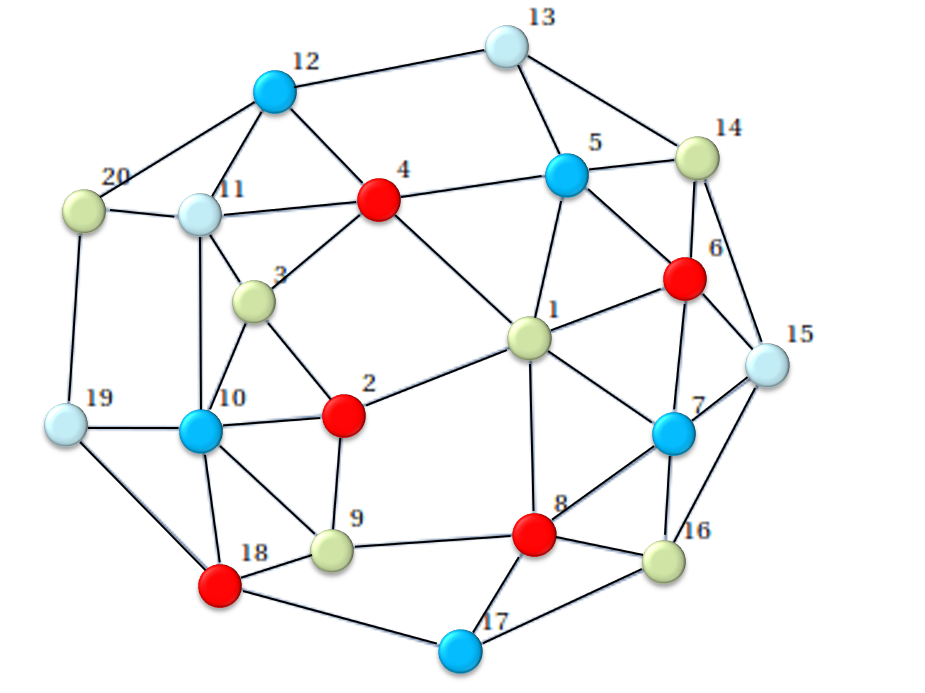}
\end{center} 
\caption{Illustration of GCP: an undirected graph (having 20 vertices) labeled using four colors.} \label{colored-graph}
\end{figure}

The Graph Coloring Problem (GCP) stands as a challenging NP-Hard combinatorial optimization problem in graph theory \citep{Jensen, Pardalos,Matula,Cheeseman,Tutte}. The GCP, namely vertex coloring problem, is a labeling problem and can be defined as minimizing the number of colors (the chromatic number) of vertices such that no two neighbouring vertices have the same color in an undirected graph. (Figure \ref{colored-graph} illustrates an undirected graph that satisfies the constraints of the GCP, with 20 vertices and painted with four colors). There are many application areas of the GCP in the real world such as optimization problems of timetabling \citep{Burke44}, scheduling \citep{Gamache}, computer science \citep{Ahmed}, biology \citep{Liu24}, sociology and etc. 

\begin{figure}[t]
\begin{center}  
\includegraphics[width=0.8\textwidth]{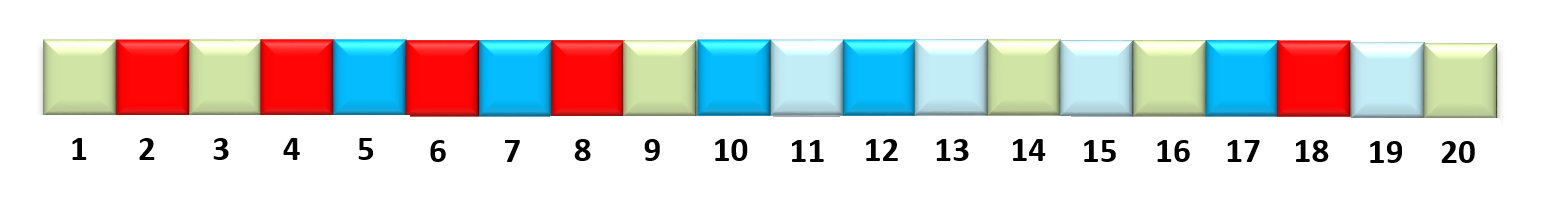}
\end{center} 
\caption{The representation of the GCP solution in Figure \ref{colored-graph}. The indices represent vertex IDs in the graph.} \label{chromosome}
\end{figure}

%The GCP is an intractable problem. 
Finding an optimal solution for the GCP is still challenging due to the time complexity of the exact algorithms that can only find solutions for graphs with vertices up to 100. Often, more intelligent techniques such as metaheuristics need to be employed for larger problems. Many metaheuristic algorithms have been proposed to obtain near-optimal solutions for the GCP recently \citep{Mostafaie}. Mostafaie et al. presented a comprehensive survey of the most influential studies in this domain. Genetic algorithms (GA), Artificial Bee Colony Optimization (ABC), Ant Colony Optimization (ACO), Cuckoo Optimization Algorithm (COA), and Particle Swarm Optimization (PSO) are the most well-known metaheuristics applied to the solution of the GCP. 

%a paragraph on metaheuristics
In essence, all metaheuristics are heuristic methods designed to solve generic combinatorial optimization problems. Often, metaheuristic approaches aim to find solutions to complex problems using a three-step process applied iteratively \citep{Falkenauer}. The first step of this three-step process involves initialization of the metaheuristic with a set of random or semi-random solutions called the \textit{candidate solutions}. Depending on the metaheuristic, the representation of the candidate solutions may vary significantly. The initialization is then followed by a fitness calculation where each candidate solution is assigned a quality-score according to its resemblance to the optimization objective. Last, metaheuristics employ various selection/adaptation operations to modify the existing candidate solutions so that: the solution space is scanned sufficiently well (i.e., exploration) and regions in the solution space that have relatively high potential for an optimal solution are searched well (i.e., exploitation). In the literature, many different selection/adaptation mechanisms exist. For example, GA mimics crossover and mutation mechanics in Darwinian biology, ABC and ACO adopt animal behavioral patterns, and PSO uses a simulated social behavior mechanism for the purpose. By adopting an iterative approach, metaheuristics explore the solution space for an optimal solution and this process continues until a specified criteria is satisfied. Figure \ref{chromosome} illustrates the representation of a solution (a chromosome for GA) used by metaheuristic algorithms for the graph given in Figure \ref{colored-graph}.

Ensemble learning is a machine learning technique that combines multiple models to improve the overall performance and accuracy of the solution \citep{Dietterich, Zhou22}. Instead of relying on a single model, ensemble methods use a collection of diverse models and aggregate their predictions to make more robust and reliable predictions than individual models alone. The diversity among the models can be achieved through different algorithms, variations in training data, or by tweaking the model parameters. Ensemble methods can leverage multiple learning methods to enhance the performance beyond what could be achieved by any individual learning algorithm in isolation \citep{Mitsutake, Singh}. They have evolved as a means to increase the generality of search and optimization algorithms and in our study, they work on top of a set of metaheuristics to solve a particular optimization problem. They can be used as distinct methodologies developed to solve challenging NP-Hard problems. In essence, ensemble algorithms can learn (decide) the most appropriate metaheuristic to solve a problem. The metaheuristic algorithms can obtain (near)-optimal solutions when exact algorithms face challenges due to the long execution times \citep{Boussaid, Nayyar}. According to the No Free Lunch (NFL) theorem \citep{Wolpert}, the performance of a metaheuristic for a specific problem is balanced by its performance on another class of problems. Therefore, for problems with different complexities, different metaheuristics may perform better. This means different metaheuristics may perform superior to others for different GCP. Combining multiple metaheuristics with ensemble algorithms, and using parallel computing to reduce the costs may provide a novel but intuitive approach to complex graph coloring problems with a state-of-the-art approach. This was the most important motivation for our study.

In addition to this, the fitness calculation of each candidate solution is an important computational effort in metaheuristic algorithms, especially considering that it must be performed at each iteration thousands of times. In some studies, this process lasts for hours or even days \citep{Lu2010}, which is one of the major drawbacks of metaheuristics. In this sense, parallel metaheuristic algorithms can achieve much better results with their high computational capacities \citep{Alba,Alba2, Talbi22}. However, parallelizing the fitness evaluation of each candidate solution is not straightforward due to the dependencies between candidate solutions, which can lead to communication overhead and load-balancing issues. Providing scalability and speed-up requires careful design and optimization of the parallel algorithm.

% gelilistirilen algoritmalari anlat....
In this study, we proposed a new and scalable parallel ensemble algorithm (PEM-Color) for the GCP optimization of large graphs. The proposed approach combined three state-of-the-art metaheuristics to achieve the best possible result quality for a given GCP. To this end, we carried out this study by using the most effective and up-to-date metaheuristics, HHO \citep{Heidari}, ABC \citep{Karaboga}, and TLBO \citep{Rao22}, with a new approach that we have not encountered in this field before. We use high-performance parallel computing techniques and Message Passing Interfaces (MPI) libraries to facilitate distributed memory computation \cite{Walker} and report significant improvements on the well-known Center for the Discrete Mathematics and Theoretical Computer Science (DIMACS) benchmark graph problem instances \citep{Ostergard, Johnson}. We report the best results in 37 out of 43 and for the rest of the problems, our results are only marginally different than the reported best solutions.

The contributions of our study are as follows:

\begin{itemize} 
    \item A new parallel ensemble metaheuristic algorithm is proposed for the GCP for the first time in literature.
    \item HHO, ABC, and TLBO are combined for the first time in an ensemble metaheuristic algorithm for the GCP.
    \item HHO is applied to the GCP for the first time.
    \item Almost a linear speed-up and strong scalability are achieved with the new parallel algorithm.
    \item Stagnation is controlled with different seeding mechanisms for each population in the computing environment.
    \item The execution time of the metaheuristics is significantly reduced using MPI libraries.
\end{itemize}

Related studies for solving the GCP are given in Section \ref{section:related}. The proposed parallel ensemble metaheuristic algorithm is presented in section \ref{section:proposed-algorithm}. The experimental setup and obtained results are reported in section \ref{sec:exp}. The conclusion and future works are provided in the last section.

%\clearpage
\section{Related work} \label{section:related}

Within this section, we conducted an extensive review of the most effective exact, metaheuristic, ensemble, parallel, tabu search, and parallel local search algorithms proposed for addressing the GCP.

\cite{Mostafaie} prepared an inclusive report, and detailed review of nature-inspired metaheuristics for the GCP.  This study emphasizes the best algorithms developed for GCP. The advantages and disadvantages of the algorithms are examined, and the key issues are discussed. We briefly mention some of the recent metaheuristics in this study.  \cite{Dorrigiv} developed an Artificial Bee Colony (ABC) algorithm \cite{Karaboga} to solve the GCP. A sequence of nodes is produced, and a color assignment algorithm is generated with this algorithm. Recursive Largest First (RLF) is used in the algorithm. \cite{Bessedik} presented ACO \citep{Dorigoaco} for the GCP. Construction and improvement strategies are used in the algorithm. Recursive Largest First (RLF) and tabu search are employed for the strategies. Experiments are performed with DIMACS graphs. The algorithm outperforms hybrid algorithms on large instances. \cite{Djelloul} developed a discrete Bat Algorithm \cite{Yang} for the GCP. The algorithm is based on the echolocation behaviour of bats. The experiments on DIMACS benchmark graphs verified the effectiveness of the algorithm. \cite{Mahmoudi} developed Cuckoo Optimization Algorithm (COA) for the GCP. Arithmetic operators (addition, subtraction, and multiplication) are used in the algorithm. After comparisons with heuristic methods, results confirmed the performance of the proposed algorithm.  \cite{Agrawal} developed a Particle Swarm Optimization (PSO) \citep{Kennedy} algorithm for the GCP. A new neighbourhood search operator is used to particles to improve the solutions. The results revealed that better near-optimal solutions are obtained for benchmark graphs.  \cite{Fleurent} developed the first applications of GA, and some hybrid algorithms for the GCP in 1996. 

\cite{Bravyi} applied the recursive quantum optimization algorithm (RQAOA) to the k-vertex graph colouring problem. The algorithm is compared to the classical and hybrid quantum algorithms. They reported that the RQAOA is surprisingly competitive.   \cite{Gaspers} presented a polynomial algorithm to compute the number of independent sets in a graph. This algorithm is a fast one for the GCP.  \cite{Shimizu} presented a polynomial quantum algorithm using quantum random access memory. The algorithm is based on quantum dynamic programming.  \cite{Hoeve} introduced an iterative framework using decision diagrams for the GCP. The decision diagram represents color classes. A constrained minimum network flow model is used to compute conflicts. They performed evaluations on 137 DIMACS graph coloring instances. They solved 52 of the instances optimally.  \cite{Sabar} investigated a new hyper-heuristic algorithm for solving examination timetabling problems, where they use coloring heuristics such as largest, saturation, largest colored degree, and largest enrollment. 

% parallel algorithms
We can briefly summarize the prominent parallel metaheuristic algorithms in the literature.  \cite{Dokeroglucolor} proposed a memetic TLBO algorithm (TLBO-Color) combining with a  tabu search. A parallel TLBO-Color is developed for painting large graphs having millions of edges and thousands of vertices. The optimization times are polynomial, and most of the optimal results of the DIMACS graphs are reported. \cite{Gebremedhin5} developed a shared memory parallel algorithm. They suggested a new ordering of the vertices.  \cite{Kokosinski} proposed a parallel GA (PGA) for computing the chromatic upper bounds.  \cite{Hijazi} introduced a parallel heterogeneous ensemble feature selection method incorporating three metaheuristics (genetic, PSO, and grey wolf optimizer). The approach involves the phases—distribution, parallel ensemble feature selection, combining and aggregation, and testing. A sequential approach on CPU, a parallel approach on a multi-core CPU, and a parallel approach utilizing both multi-core CPU and GPU are implemented. The proposed parallel approach significantly enhanced predictive outcomes and reduced running time. This is the study most similar to our proposed algorithm. We use MPI instead of GPU parallel computation environment, which is more scalable.

\cite{Sariyuce} investigated the application of Largest First and Smallest Last vertex-visit orderings. They studied a distributed post-processing operation on large graphs. They developed multi-core architectures for distributed graph coloring algorithms.  \cite{Lu} achieved a balanced coloring of a graph. They propose multiple heuristics with parallelization approaches for multi-cores. They studied the impact of the proposed balanced coloring heuristics.  \cite{Huang4} showed that deep reinforcement learning can effectively paint large graphs. \cite{Jones} presented an asynchronous graph coloring heuristic for parallel computers. Experimental results demonstrated that the heuristic is scalable and can use three or four more colors than the best heuristics. Execution time improvements are also observed in this new algorithm. \cite{Gebremedhin} presented a fast parallel heuristic for shared memory programming with linear speedup. The heuristics are developed using OpenMP \citep{Chandra}. Experiments validated the theoretical run-time analysis of the proposed algorithm.  \cite{Hasenplaugh} introduced the ordering heuristics, smallest-log-degree-last (SLL), and largest-log-degree-first (LLF) for parallel greedy algorithms. They showed that LLF and SLL allow for speedups.  \cite{Gjertsen} introduced a new parallel heuristics.  Experimental results showed that the heuristics effectively produce a balanced colouring and exhibit scalability.  \cite{Allwright} introduced new parallel graph coloring algorithms using sequential heuristics. The algorithms are developed on Single Instruction, Multiple Data, and Multiple Instruction, Multiple Data architectures. \cite{Lewandowski} introduced a new hybrid algorithm combining a parallel version of Morgenstern's S-Impasse exhaustive search algorithm \citep{Morgenstern}. The algorithm works well without tuning the parameters of the algorithm.

\cite{Naumov} studied parallel graph coloring algorithms. They implemented different heuristics and verified their performance on Graphics Processing Units (GPU). They presented numerical experiments that showed the performance of the algorithms. A speedup of almost 8 times on the GPU over the CPU is achieved.  \cite{Abbasian} proposed PGA to solve the GCP. A new operator is developed to improve the solution quality. This operator produces high-performance individuals and inputs them into the population. The experiments verified the performance of the new algorithm on the DIMACS graphs.  \cite{Chen} presented a parallel graph coloring implementation on GPUs. They used the speculative greedy scheme.  Experimental results showed that the proposed algorithm achieves up to 8.9 times speedup over its serial version.  \cite{Alabandishort} described two improvements for the Largest Density First heuristic. They presented a short-cutting approach to increase the parallelism by color reduction techniques. The algorithm yields 2.5 times speed-up.

\cite{Catalyurek} proposed a set of multi-threaded heuristics for the GCP. The algorithms are developed on shared memory and rely on iterations and speculation. Cray XMT, Sun Niagara 2, and Intel Nehalem are used during the experiments. The experiments verified the performance of the algorithms on multi-core architectures for irregular problems.  \cite{Bogle} presented new MPI, and GPU coloring techniques on the distributed memory architectures. The algorithms use Kokkos Kernels for CPUs and GPUs. They extended the algorithm to compute distance-2 coloring problems. They proposed a novel heuristic to reduce communication.  \cite{Alabandi} presented an approach that increases the parallelism, but not the quality of coloring. The technique is observed to yield 3.4 times speed-up. The CUDA version on a Titan V is observed to be 2.9 times faster.  \cite{Grosset} performed experiments on the features of GCP algorithms implemented on  GPUs. They reported that many cores and high global memory bandwidth need distinct approaches for parallel computation. Their algorithm solves coloring conflicts on the GPU by iterating. \cite{Boman} described a new parallel algorithm for distributed memory. The algorithm works iteratively. At each iteration, vertices are colored, and a set of incorrectly colored vertices is identified. Speedup is provided by reducing the communication between processors.  \cite{Giannoula} proposed the ColorTM algorithm that detects inconsistencies of coloring between vertices. ColorTM algorithm has a speculative synchronization to minimize costs and improve parallelism. \cite{Deveci} proposed an edge-based approach that is suited to GPUs. They verified that GPUs reduce the chromatic number significantly.  \cite{Osama} 
 designed parallel algorithms on the GPU using data-centric abstractions. They analyzed the variations of a baseline independent-set algorithm on solution quality and execution time. Their implementation produces 1.9 times fewer colors than the other recent implementations.

\cite{Thadani} proposed applications of graph coloring for team-building, scheduling, and network analysis problems. \cite{Kole} compared ACO, simulated annealing, and quantum annealing for the GCP. Datta et al. proposed a new graph coloring algorithm for scheduling examinations for universities and colleges across \cite{Datta}.   \cite{Ananda} developed a tabu search graph coloring algorithm for the scheduling problem of nurses in a hospital.  \cite{Goudet} developed a new framework of deep neural networks with heuristics for the GCP. The algorithm was able to obtain competitive results.

\cite{Barenboim}  published a book about distributed GCP and its fundamental theories and recent studies.  \cite{Xuswarm} proposed a distribution evolutionary algorithm (DEA-PPM). The DEA-PPM algorithm employs a novel probability model to search the distribution space. A tabu search method is used in this study. The algorithm leads to a nice balance between exploration and exploitation.  \cite{Freitas} consider coloring problems with distance constraints as weighted edges and modeling them as distance geometry problems (DGPs). The authors proposed new variations of vertex coloring problems in graphs.  \cite{Seker} studied the NP-Hard selective GCP. The authors proposed an algorithm to generate perfect graphs and produce a collection of instances. The experiments demonstrate that their proposed solution improves the solvability of the problem.

Despite the substantial progress made in the field of graph coloring, there remain several intriguing research gaps that warrant further exploration and investigation. Large instances of the GCP are still challenging and need robust approximation algorithms. Dynamic graph coloring, multi-objective graph coloring (minimizing both chromatic number and execution time), parallel and distributed graph coloring to improve the efficiency of graph coloring using modern advanced computing platforms, machine learning integration to this optimization problem, robustness and resilience, and real-world applications, bridging the gap between theoretical advancements and practical applications are crucial issues. Research that focuses on tailoring graph coloring algorithms to specific real-world problems, such as scheduling, resource allocation, or frequency assignment, is essential. Addressing these research gaps will not only contribute to a deeper understanding of the GCP but also enhance the practical applicability of graph coloring algorithms across various domains.

To our knowledge, our proposed algorithm is the first study to develop a parallel ensemble metaheuristic algorithm for the GCP using state-of-the-art metaheuristics, HHO, ABC, and TLBO, and to demonstrate its effectiveness on large graph instances from the well-known DIMACS benchmark.

%\clearpage
\section{PEM-Color Framework}
\label{section:proposed-algorithm}

\begin{figure}[t]
\begin{center}  
\includegraphics[width=0.8\textwidth]{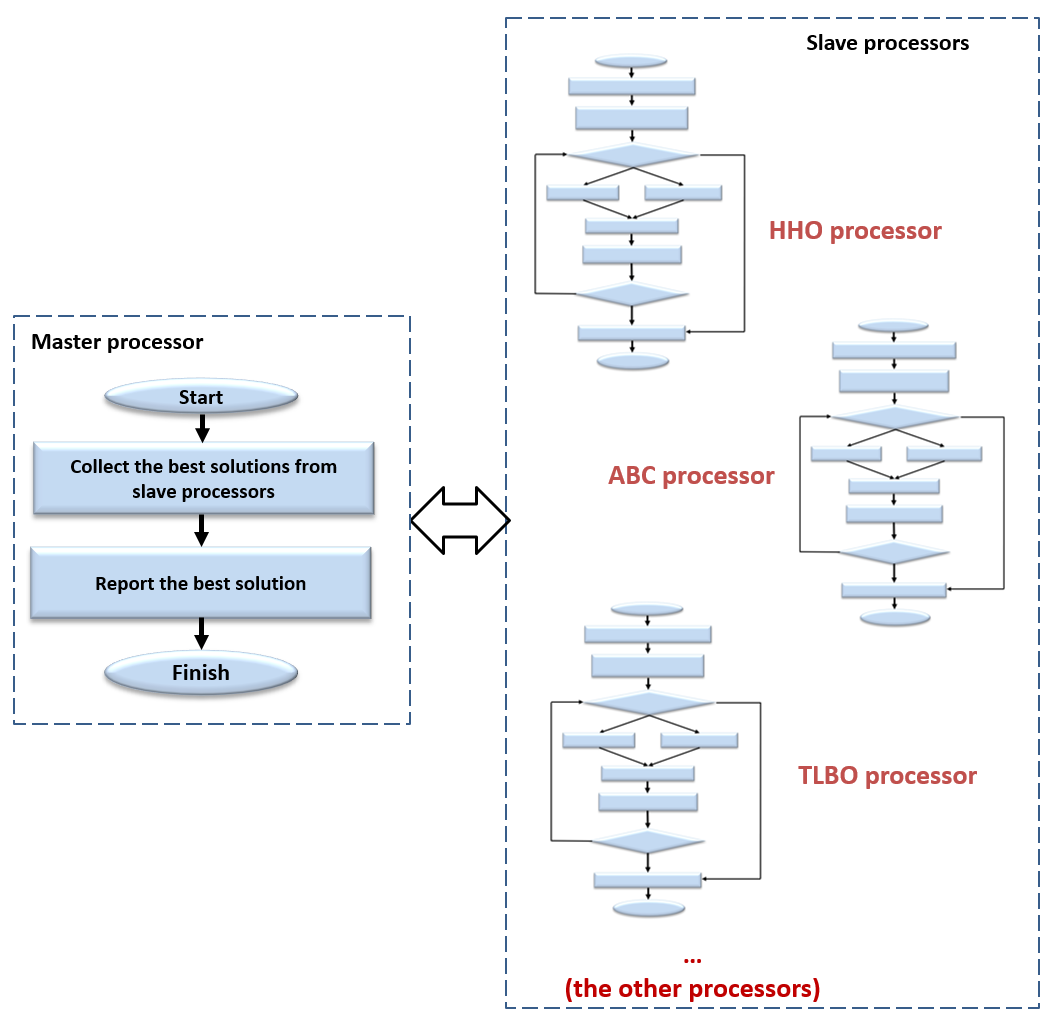}
\end{center} 
\caption{The parallel computation framework of PEM-Color algorithm. Master and slave nodes and how they interact is presented in this drawing.} \label{fig:flowchart-parallel}
\end{figure}

This section describes our proposed parallel ensemble metaheuristic algorithm for the GCP (PEM-Color) and provides brief information about the metaheuristics (HHO, TLBO, ABC) and the local search technique (TabuCol) used for the development of the PEM-Color algorithm. The ensemble algorithms can leverage the strengths of various metaheuristic algorithms, compensate for their weaknesses, and provide more accurate and robust solutions for the GCP. This approach, amenable to also facilitate high-performance parallel computation techniques, is applied for the first time in this field to the best of our knowledge. The PEM-Color algorithm is an island parallel ensemble algorithm that distributes the optimization process of multiple metaheuristics (HHO, ABC, and TLBO) evenly among many processors, with one processor allocated to each metaheuristic. For example, if there are 64 processors, one processor serves as the master node, and the remaining 63 processors are allocated to the three different metaheuristics (21 processors for each metaheuristic). The algorithm also uses the TabuCol algorithm as a local search algorithm during the exploitation phase of each candidate solution in the population of the metaheuristics. The initial individuals of each population are diversified to provide a better exploration possibility, and in our experiments, the population size is set to 20 for each metaheuristic. The PEM-Color algorithm is implemented using MPI libraries and communication between processors is minimized to provide scalability. For synchronization purposes, the slave nodes send their best solution in the population to the master node only at the end of the generations. The master node then reports the overall best solution. The implementation and combination of the metaheuristics are explained in the following subsections. The flowchart and pseudocode of the parallel PEM-Color algorithm are illustrated in Figure \ref{fig:flowchart-parallel} and Algorithm  \ref{algorithm:PEM-Color}, respectively.

\begin{algorithm}
\caption{Pseudocode of the Island Parallel Ensemble PEM-Color algorithm.}\label{algorithm:PEM-Color}
\begin{algorithmic}[1]

\State Input Graph $G$($V$, $E$) where $V$ is the set of vertices and
\State $E$  : set of edges;
\State
\If{(I am the \textbf{master} node)}

    %\While{($i$ is smaller than $T$)}
    \While{(i++ $<$  \#slaves)}
    %\While{(i++ < \#slaves)} do
      \State Receive the best local solution from slave\_i();
      \State Update the best solution();
    \EndWhile
    
\State Report the overall best solution ();
\EndIf
\State
\If{(I am \textbf{slave})}
\State  Produce\_initial\_population();
\State  Calculate\_fitness\_values();
\State
     \State Apply the metaheuristic(); // HHO, ABC, or TLBO
     \While{(i++ $<$  \#generations)}
     \State Exploration Step
     \State Exploitation Step using TabuCol();
     \EndWhile
\State
\State Send the best solution to the master node();
\EndIf		

\end{algorithmic}
\end{algorithm}

\subsection{Parameter tuning of the metaheuristics}

The parameters of the metaheuristics are randomly selected within the defined ranges, making the parameter setting at each processor unique. Therefore, we provide a comprehensive method depending on the type of graphs that we paint. Various settings are available when parallel versions of the ABC, HHO, and TLBO are executed concurrently. 

The ABC algorithm exhibits simplicity and adaptability in its parameter settings. Key parameters include the colony size, representing the number of artificial bees, and the maximum cycle count, determining the algorithm's convergence criteria. Additionally, ABC features exploration/exploitation factors, regulating the exploration-exploitation trade-off. Fine-tuning these factors influences the balance between global and local search capabilities. While the algorithm is relatively robust to parameter changes, employed bee, and onlooker bee ratios are adapted to suit specific graph characteristics. The effectiveness of the ABC algorithm stems from its ability to strike a balance through minimal yet impactful parameter adjustments.

The key parameters of the HHO algorithm include the population size, representing the number of hawks in the search space, and the convergence factor, controlling the algorithm's termination criteria. Additionally, HHO employs the exploration factor to strike a balance between global and local search capabilities, enhancing adaptability across different problem landscapes. The attack strategy parameter guides the hawks' hunting behavior, influencing the optimization process. Our parallel algorithm can fine-tune these parameters to suit specific graph coloring optimization challenges, allowing HHO to efficiently converge towards optimal solutions while maintaining robustness across diverse problem domains.

TLBO is renowned for its simplicity, requiring no manual parameter tuning. This metaheuristic algorithm relies on a collaborative teaching and learning approach inspired by the classroom environment, making it inherently parameter-free. TLBO's self-adaptive nature enhances its appeal, eliminating the need for user intervention in parameter adjustments.

%\clearpage
%\input{03_01_HHO.tex}
\subsection{Harris Hawk Optimization Metaheuristic for the GCP}

\begin{figure}[t]
\begin{center}  
\includegraphics[width=0.5\textwidth]{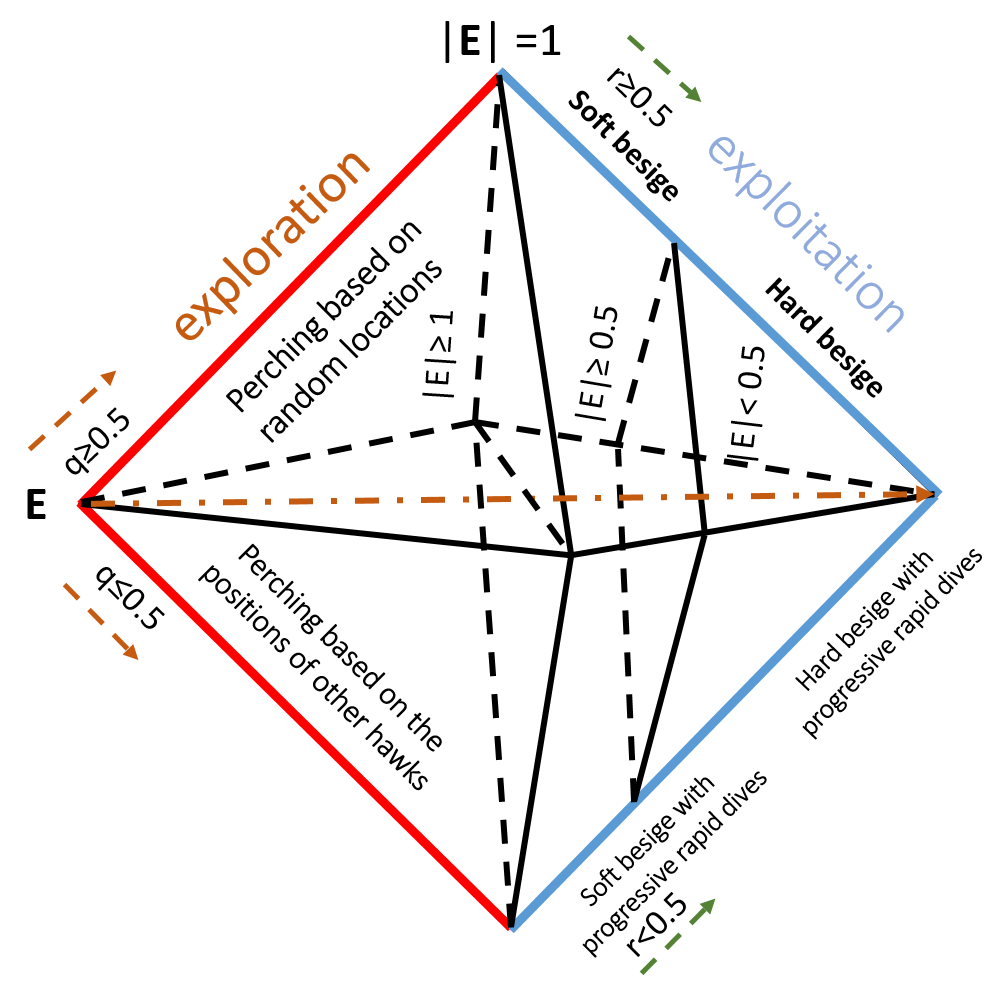}
\end{center} 
\caption{Harris Hawk (HH) metaheuristic hunting strategy summary. The HH metaheuristic exhibits distinct stages that depend on the energy levels of the hawks. These stages correspond to various behaviors and strategies adopted by the hawks, which are influenced by their energy conditions. \label{figure1-HHO}}
\end{figure}

HHO is one of the metaheuristics of the PEM-Color algorithm. The HHO metaheuristic simulates the hawks that attack prey from various directions synchronously \citep{Heidari, Dokerogluisland}. These birds use some patterns to exhaust their prey. The HHO is gradient-free, and inspired by the attacking strategies of the hawks. Figure~\ref{figure1-HHO} shows the steps of the HHO according to the energy level, $E$, and the parameter, $q$. In algorithm \ref{alg:HHO}, the details of the HHO algorithm that uses TabuCol, as a local search algorithm are presented. 

\begin{algorithm} [t]
	\caption{Pseudocode of the HHO algorithm.}
	\label{alg:HHO}
\begin{algorithmic}[1]

\State  $N$: is the population of hawks
\State  $T$: \# of iterations 

Initialize $X_{i}(i=1,2,\ldots,N)$

	\While{($i$ $<=$ $T$)}
		\State Calculate the fitness values	
		\State Update the hawks into the population 			
		\State Set \textbf{$X_{rabbit}$} as rabbit's location
		
		\For{(each hawk ($X_{i}$))}
        \State Update  $E_{0}$ 		
		    \State  Update the value of $E$ 
    
		    \If{($|E|$ $>$ 1)}   		
		      \State Perform Exploration
		    \EndIf 
										
	 \If {($|E|$ $<$ 1)} 		 
  	\State  Perform Exploitation
	 \EndIf			
	\EndFor
			\State	TabuCol(hawk($X_{i}$));			
  \EndWhile		
	\State \textbf{Return} \textbf{$X_{rabbit}$} // the best solution 
	
\end{algorithmic}
\end{algorithm}

In their quest to find prey, the hawks employ a perching strategy, often perched on trees while observing the hunting field. The perching behavior is guided by the positions of other hawks and the prey, but this guidance only occurs when the value of $q$ is less than 0.5. Otherwise, the hawks resort to random perching. This study introduces two exploration operators. The first one, exploration\_1, involves selecting two hawks, hawk\_1 and hawk\_2, and transferring a subset of features from hawk\_1 to hawk\_2. The second operator, exploration\_2, copies selected features from the best-performing hawk to the current hawk. Subsequently, the newly generated candidate is incorporated into the population. The determination between the exploration and exploitation phases relies on the prey's escaping energy value ($E$), which follows a decreasing trend as outlined below: 

\begin{equation} \label{equ:energy}
E=2E_{0}(1-\frac{t}{T})
\end{equation}

In the context of the algorithm, $E_{0}$ represents the initial energy, and $T$ denotes the total number of iterations. The initial energy $E_{0}$ lies within the range of (-1, 1) and gradually increases from 0 to 1 as the prey becomes stronger. Conversely, if $E_{0}$ decreases from 0 to -1, it indicates a decline in the prey's mobility. As the number of iterations progresses, the value of $E$ decreases. When $E \geq 1$, the HHO algorithm leans towards exploration; otherwise, it focuses on exploitation.

The hawks employ a besieging strategy around the prey, contingent upon the rabbit's energy level. If $ E \geq$0.5, a soft besiege is executed; otherwise, a hard besiege is employed. In instances where both $r\geq0.5$ and $E \geq0.5$, signifying a good energy level in the rabbit, it can execute random jumps to elude the hawks (where $r$ is a random value). Meanwhile, the hawks can encircle the prey and execute surprise jumps. These surprise jumps are initiated by generating a random number ($J$), and the features of the rabbit are then copied into the current solution. Prey performs random jumps if $r\geq$ 0.5 and $ E <$ 0.5.The position of the hawk changes as described by the following equation: 

\begin{equation} \label{equ:hard}
 X(t+1)=X_{rabbit}(t)-E \left \vert \Delta X(t) \right \vert
\end{equation}

\noindent At iteration $t$, $\Delta X(t)$ represents the disparity between the hawk and the current rabbit. In this context, a specific dimension of the prey is copied. To fend off attacks, the rabbit can nullify them when $r < 0.5$ and $E \geq 0.5$. A novel operator, characterized by a high perturbation value, has been introduced to facilitate this action. Depending on the prey's energy level, a subset of features is extracted and then copied to the most recent hawk.\\

In scenarios where $E <$ 0.5 and $r < $ 0.5, the hawks are presumed to initiate attacks to minimize the distance. To achieve this, certain features are selected from the prey and transferred to a randomly chosen hawk. The number of features chosen is kept minimal to avoid excessive perturbations. For a detailed implementation, refer to the pseudocode of the HHO metaheuristic presented in Algorithm~\ref{alg:HHO}.

%\clearpage
%\input{03_02_ABC.tex}
\subsection{Artificial Bee Colony Optimization Metaheuristic for the GCP}

The ABC metaheuristic is inspired by the behaviour of bee colonies \citep{Karaboga}. The ABC metaheuristic makes use of exploring and exploiting the resources of food (solutions). The hive has three types of bees: employed, observer, and scout. The employed search for food sources and discuss their findings by dancing. When a worker has finished collecting nectar, it transforms into a scout and searches for alternative sources. Onlookers observe how the hired bees dance and select food. Scouts look for new sources. In the first step, the initial population is produced randomly. This optimization procedure is repeated with a new group of bees. A forager begins as an unemployed bee without any knowledge. A regular bee can act as a scout, exploring the solution space or observing the other bees, looking for new food sources. The bee gets the food and collects the nectar in the hive. Scout bees are randomly deployed to search the issue space during the initial phase of each iteration and return to discuss their findings with other bees. Onlookers choose the top candidates for exploitation. The employed flies to the examined solutions and the exploitation begins according to the parameters of the ABC. After the iterations are completed, the best result is provided as the solution. 50\% of the population is the number of onlooker bees, the other is the number of employed bees, and there is a single scout bee in the hive population. The pseudocode of the metaheuristic is given in  Algorithm \ref{algorithm:ABC}. 

\begin{algorithm} [t]
\begin{algorithmic}[1]
	\caption{Pseudocode of the ABC algorithm.}
	\label{algorithm:ABC}

	\While{(i++ $<$  \#max\_iterations)}
	    \indent Scouts forage for food();
             \State Scouts return and dance();
             \State Onlookers evaluate the sources();
             \State Checking visited resources();
             \State Choose the best resources();
             \State Employed bees go to sources();               
             \State Collect the food ();	
         \EndWhile	
  
\end{algorithmic}     
\end{algorithm}

%\clearpage
%\input{03_03_TLBO.tex}

\subsection{Teaching Learning-Based Optimization Metaheuristic for the GCP}

The learning environment of the students inspires the TLBO in a classroom  \citep{Rao22, Dokeroglucolor}. The knowledge of individuals is improved by the teachers and classmates. Teaching and Learning are the two phases of this metaheuristic. There exist no specific parameters in the TLBO algorithm. The iterations of the TLBO continue until the minimum chromatic $k$ value is obtained or the number of generations is finished. Algorithm \ref{alg:TLBO} shows the pseudocode of the TLBO metaheuristic. Each gene in the chromosome is a vertex in the solution structure of the GCP. The partition crossover operator used in this algorithm generates new diversified individuals. The initial population is generated using random solutions.

\begin{algorithm} [t]
\begin{algorithmic}[1]
	\caption{Pseudocode of the TLBO algorithm.}
	\label{alg:TLBO}
\State \textbf{Input:} $G$ is a graph, $k$ is the number of colors to be used,\\
 $P$ is the population. 
\State \textbf{Output:} The best solution, $s^*$, obtained by the algorithm.\\

Generate\_initial\_population$()$;
Calculate\_fitness\_values($P$);

\While{$i++ < \#generations$ }
  \State $p_1, p_2 \leftarrow P$;
  \State // learning phase
  \State  $s_0$ $\leftarrow$ Crossover ($p_1, p_2$);
  \State  $s_0$ $\leftarrow$ TabuCol($s_0$);
  \State  Insert\_into\_population($S_0, S_{1}, ....., S_{|P|}$)
\EndWhile

\State\textit{\textbf{return}}  the\_best\_solution($s^*$);

\end{algorithmic}     
\end{algorithm}

%\clearpage
%\input{03_04_TabuCol.tex}

\subsection{Tabu Search Algorithm, TabuCol}

The TabuCol \citep{Hertz123, Galiniertabucol} is an efficient local search algorithm proposed in 1987. It starts with an initial solution ($s$) and tries to improve this solution iteratively. The neighbours of the solution $s$ are defined as $N (s)$. Candidate solutions are generated in $N(s)$. The search is directed to $s^*$ when $f(s^*)$ has a better value than $f(s)$. The stopping condition of TabuCol is when \textit{f* = 0}. A random node \textit{x} is selected among the neighbors and assuming $x \in V_i$, a random color $i \neq j$ is selected and $s'$ is obtained from $s =( V_1,...,V_k$) as given below:

\begin{equation}
V'_j = V_j \cup \{x\}; V'_i = V_i \backslash \{x\}; V'_r = V_r \; for \; r=l,...,k; r \neq j, r
\end{equation}

Tabu list ($T$) is the most important component of TabuCol. Repeated moves are prevented using this list. For each move ($s \rightarrow s^*$), the opposite move ($s^* \rightarrow s$) is inserted to the end of $T$, and the oldest one is deleted. The TabuCol algorithm stops after $nbmax$ many iterations. When a vertex \textit{x} is moved from $V_i$ to $V_j$, the pair \textit{(x, i)} becomes tabu. An aspiration value \textit{A(z)} is used to follow the aspiration level, \textit{z = f(s)}. If a move is a tabu, but $f(s') produces \leq A (z)$, then the status of the move is not tabu anymore. The tabu list size ($\vert T \vert$) is another crucial parameter. It is assumed to be seven during our experiments. With small tabu list sizes, cyclings may occur, whereas larger values may increase the computation time. Detailed pseudocode of TabuCol can be seen in Algorithm \ref{algorithm:tabucol}.

\begin{algorithm} [t]
\caption{Pseudocode of the TabuCol algorithm.}\label{algorithm:TaBUcOL}\label{algorithm:tabucol}
\begin{algorithmic}[1]
\State $G$=($V$, $E$), $k$ is the \# of colors
\State $T$: the tabu list
\State $rep$: the number of neighbouring solutions of $s$
\State $nbmax$: the maximum \# of iterations

\State $V_{1}, ....., V_{k}$ are the set of colors. 

\State Produce an initial solution $s$= ($V_{1}, ....., V_{k}$);
\State count = 0; 
\State Produce a tabu list $T$. 
 
\While{(f(s) $>$ 0 and count $< nbmax$)}

Produce $rep$ many neighbours $s_i$ of \textit{s} with 

move $s \rightarrow$ $s_i \notin T$ or f($s_i$) $\leq$ A (f(s));

$s'$ = best neighbour;

Update  $T$;

Move $s$ $\leftarrow$ $s_i$  and delete the oldest move from $T$;

$s = s'$ ;

count ++ ;
\EndWhile

\State If $f (s)$=zero, $G$ is colored with $k$ colors 
\State Otherwise no solution with $k$ colors is obtained 

\end{algorithmic}
\end{algorithm}

%\clearpage
\section{Evaluation of experimental results and discussion} \label{sec:exp}

In this section, we give details of the experimental setup, graph instances of DIMACS, the results, and the comparison of the PEM-Color algorithm to the state-of-the-art algorithms in the literature.

The experiments were conducted on an AMD Opteron Processor 6376, featuring a Non-Uniform Memory Access (NUMA) multiprocessing architecture. The processor boasts 4 sockets, each equipped with eight cores, and each core can support two threads. In our setup, each node is equipped with eight processors, which collectively share a Level Cache of 6 MB. To complement the processing power, the system is equipped with 64 GB of RAM, which is distributed evenly across the eight nodes. The frequency of the clock is 1400 MHz. The DIMACS problem instances are used in our experiments. The instances are flat, Leighton, random geometric,  C2000.5 and C4000.5, and latin\_square\_10 graphs. In the library of these benchmark problems, (near)-optimal solutions are reported. In this study, we actively pursued the potential for discovering new and improved solutions for these problem instances. Our efforts focused on maximizing the number of instances where better solutions could be attained.

The population size of the metaheuristics used in our study is 20, and the number of generations is 1000. The PEM-Color algorithm collects the best results obtained from the studies. The reported results are the average value of 20 executions. The depth of the TabuCol is selected as 100,000 and the size of the tabu list $|T|$ is 7. The other random values in each metaheuristic are generated according to the processor id it is executed by. The seeding mechanism of the random values is initialized with these numbers, and it is observed to provide a good diversification and exploration/exploitation mechanism during the execution of the metaheuristics. Table \ref{parameters} gives the details of the parameters used in our algorithms. 

\begin{table}[t]
	\caption{The parameters of the metahueristics used in PEM-Color algorithm.}	\label{parameters}
	%\begin{center}
	%\scalebox{0.95}{
		\begin{tabular}{lr}  
\hline
parameter & value \\ 
\hline
population size & 20  \\ 
initial population generation & random \\ 
\# cores & 64  \\ 
\# generations & 1,000 \\ 
$E_{0}$, initial energy of HHO &  range of (-1, 1) \\ 
surprise jumps of HHO, $J$  & random \\ 
number of onlooker bees, ABC  &  50\% of the population\\ 
number of scout bees     & 1 \\ 
number of teachers   & 1 \\ 
deployment of scout bees  & random \\ 
tabu list ($T$) size &  7 \\ 
depth of TabuCol & 100,000 \\ 
\hline
		\end{tabular}
		%}
   %\end{center}
\end{table}

\subsection{The performance of PEM-Color algorithm on small problem instances }

In this part, we report the performance of the PEM-Color algorithm (using 64 processors) on small instances of the DIMACS benchmark given in Table \ref{easy-instances}. $\#vertices$ is the number of vertices, $\#edges$ is the number of edges, $k^*$ is the best-known chromatic number, $k$ is the number of colors reported by the PEM-Color algorithm, \#hits/20 is the number of times the PEM-Color algorithm hits the best solution in 20 executions, and time (sec.) is the average execution time. The best-known solutions to this set's problem instances are obtained with practical execution times in our experiments. The average execution time is 0.939 seconds. DSJC250.5 problem instance (having 0.50 edge density) has the longest execution time at 10.2 seconds. R1000.1 had the largest number of vertices in this set. However, obtaining its best solution in 0.041 seconds was very easy. A direct correspondence is observed between the execution time and the density of edges in the graphs. If the density is higher and the number of vertices is large, the execution significantly increases in our proposed algorithm. Consequently, the PEM-Color has been proven to be very successful on small DIMACS problem set graph instances.

\begin{table}[t]
	\caption{The results of the PEM-Color algorithm for 19 small DIMACS problem instances.}	\label{easy-instances}
	%\begin{center}
	%\scalebox{0.95}{
		\begin{tabular}{lrrrrcr}  
\hline
Instance & \#vertices & \#edges  & $k^*$ & $k$& \#hits/runs  & time(sec.)\\ 
\hline
DSJC125.1 &125 &736  &5&5 &20/20 & 0.026 \\ 
DSJC125.5 &125 &3891 &17&17 &20/20&0.135 \\ 
DSJC125.9 &125 &6961  &44&44 &20/20&0.011 \\ 
DSJC250.1 &250 &3218  &8&8 &20/20&0.018 \\ 
DSJC250.5 &250 &15668  &28&28 &20/20&10.214 \\ 
DSJC250.9 &250 &27897  &72&72 &20/20&1.822 \\
DSJR500.1 &500 &3555   &12 &12 &20/20&0.023 \\
school1 &385 &19095  &14 &14&20/20&0.098 \\
school1\_nsh &352 &14612  &14 &14 &20/20&0.070 \\
flat300\_20\_0 &300 &21375  &20 &20 &20/20&0.020 \\
le450\_15a &450 &8168  &15 &15&20/20&0.187 \\
le450\_15b &450 &8169  &15 &15 &20/20&0.101 \\
le450\_25a &450 &8260  &25 &25&20/20&0.007 \\
le450\_25b &450 &8263  &25 &25 &20/20&0.011 \\
R1000.1  &1000 &14348  &20 &20 &20/20&0.038 \\
R125.1 &125 &209  &5 &5  &20/20 &0.001 \\ 
R125.1c &125 &7501  &46 &46 &20/20 &4.925 \\ 
R125.5 &125 &3838  &36 &36 &20/20 &0.143 \\ 
R250.1 &250 &867  &8 &8 &20/20 &0.003 \\ 
\hline
average & 320 &9296  & 22.58 & 22.58 &20/20 & 0.939 \\
\hline
		\end{tabular}
		%}
	%\end{center}
\end{table}

\subsection{The performance of PEM-Color algorithm on large problem instances }

Table \ref{hard-instances} gives our results with harder graph problem instances from DIMACS using 64 processors. There are 24 problem instances in this set. The average number of vertices and edges are 845 and 334,456, respectively, whereas 320 vertices and 9296 edges are on average in easier problem instances. $k^*$ is also a meaningful parameter to understand the difficulty level of the problem sets. As this value increases, we observe that the difficulty of the problem set becomes harder. $k^*$ values are 22.6 and 83.5 for easier and harder problem instances, respectively. We could not find the best solutions for six of the problems reported in the literature. For 18 problem instances, we obtained the best results. We used 2040 colors to paint all these graphs in Table \ref{easy-instances}, and the total number of best solutions was 2003. We have used 37 more colors (1.84\% more) than the total sum of the best solutions in all graphs. This value is a promising result in terms of solution quality. According to the NFL theorem, there will always be new opportunities for achieving improved solutions by employing novel metaheuristics. It is observed that different metaheuristics can be better on some classes of problems while they are not good with the rest of the combinatorial problems. This is a current research topic. However, executing many metaheuristics simultaneously in a parallel computation environment can be considered to select the best among the others. 

Setting the parameters of the metaheuristics is a critical area for the quality of the solutions \cite{Huangparameter}. In our study, we have used previously optimized values of the HHO, ABC, and TLBO metaheuristics. There will always be good opportunities in this field to obtain better solutions using better parameter settings. With the results we have obtained after the experiments, most of the best results in the literature are obtained. 

\begin{table}[t!]
\renewcommand{\arraystretch}{1.3}
\caption{The results of the PEM-Color algorithm for 24 larger (harder) DIMACS problem instances.} \label{hard-instances} \centering
%\scalebox{0.95}{
\begin{tabular}{lrrrrcr}
\hline
Instances & \#vertices &  \#edges  &  $k^*$  & $k$ & \#hits/runs  &time(sec.) \\
\hline
        C2000.5        &2000 &999,836  &153  & 148  & 20/20 &  187.6\\ 
	C4000.5        &4000 &4,000,268 &280 & 272  & 0/20 &  468.2\\ 
	latin\_sqr\_10 &900 &307,350 &98	 & 98   & 20/20 &  614.1\\  
        DSJC250.5      &250 &15,668  &28     & 28   & 20/20 & 8.7\\  
	DSJC500.1 &500 &12,458  &12          & 12  & 20/20 &  670.5\\   	
	DSJC500.5 &500 &62,624  &49          & 48  & 20/20 & 22.4\\       	
	DSJC500.9 &500 &112,437  &126        & 126 & 20/20 & 420.2\\   
	DSJC1000.1 &1000 &49,629  &20        & 20   & 20/20 & 2.7 \\ 
	DSJC1000.5 &1000 &249,826  &83       & 83   & 0/20 &  88.2  \\   
	DSJC1000.9 &1000 &449,449  &224      & 223  & 0/20 &   110.8\\  
	DSJR500.1c &500 &121,275  &85        & 85   & 20/20 &  1318.8\\   	
	DSJR500.5 &500 &58,862  &122         & 122  & 0/20 &  325.7 \\ 	
	R250.5 &250 &14,849  &65             & 65   & 20/20 &   38.4\\
	R1000.1c &1000 &485,090  &98         & 98   & 20/20 & 7.7\\ 
	R1000.5 &1000 &238,267  &234         & 240  & 0/20 &  1240.1\\
 	flat300\_26\_0 &300 &21,633  &26     & 26   & 20/20&  15.7\\ 
	flat300\_28\_0 &300 &21,695  &28     & 28   & 20/20 &  956.5 \\ 	
	flat1000\_50\_0 &1000 &245,000  &50  & 50   & 20/20 &  846.3\\    
	flat1000\_60\_0 &1000 &245,830  &60  & 60   &20/20&  1705.4\\ 
	flat1000\_76\_0 &1000 &246,708  &82  & 82   & 0/20 &  198.2\\ 
	le450\_15c &450 &16,680  &15         & 15    & 20/20 & 121.4\\ 
	le450\_15d &450 &16,750  &15         & 15    & 20/20 &  840.2\\ 
	le450\_25c &450 &17,343  &25         & 25    & 20/20& 190.5\\
	le450\_25d &450 &17,425  &25         & 25    & 20/20 &  445.8\\ 
\hline
average & 845.8 &334,456.3  & 83.5 & 85 &15/20 & 451.8 \\
\hline
\end{tabular}
%}
\end{table}

\subsection{Comparisons with state-of-the-art algorithms}

We compare our solutions with 13 state-of-the-art algorithms in the literature. The algorithms are local search methodology Variable Space Search (VSS) \cite{Hertz}, Generic Tabu Search (GenTS) algorithm \cite{Dorne}, Iterated Local Search (ILS) \cite{Chiarandini}, Foopar \citep{Blochliger}, Genetic and Tabu Search algorithm (GTS) \citep{Fleurent}, Parallel Coloration Neighborhood Search algorithm (PCNS) \cite{Morgenstern22}, Hybrid Evolutionary Algorithms (HEA) \cite{Galinier}, Adaptive Memory Algorithm (AMACOL) \cite{Galinier2008}, Heuristic Algorithm (MMT) \cite{Malaguti}, MInimal-State Processing Search algorithm (MIPS\_CLR ) \cite{Funabiki}, a hybrid evolutionary algorithm for the GCP (Evocol) \cite{Porumbel}, the memetic algorithm (MACOL) \cite{Lu2010}, and P-TLBO-Color \cite{Dokeroglucolor}.

Table \ref{comparisons-state} presents the results of the comparisons. They are obtained from the goals of other studies. When the results obtained are compared with other algorithms, it can be seen that the PEM-Color algorithm is among the state-of-the-art algorithms. PEM-Color achieved the best results in the literature for 18 out of 24 problems. With large graph instances, the algorithm has a promising performance. Adding more processors to the computation environment positively affects the performance of the PEM-Color algorithm. 

We would like to note here that, although at first sight, Evocol and MACOL algorithms performed equally well, reporting some of the best-known results for some of the datasets, these two studies concentrate on result quality as the main metric and are allowed to execute the proposed algorithms for an unlimited amount of time. The corresponding articles report execution times up to 120 hours which may be a questionable setting for larger, more practical settings. Additionally, the differing objectives of these studies disparage a direct comparison with PEM-Color. Even for the hardest problem instances, PEM-Color was able to find results in less than 28 minutes, which highlights additional merits for larger and more practical applications. 

The innovative PEM-Color algorithm, initially designed for efficiently solving large instances of the GCP, presents a promising avenue for addressing other NP-complete problems, including the Traveling Salesman Problem (TSP) and Maximum Clique Problem (MCP). While the paper briefly mentions the applicability of PEM-Color to diverse problem domains, further exploration of its potential impact on TSP and MCP, supported by preliminary results or theoretical discussions, would elucidate the broader significance of this research. Ensemble algorithms, known for their effectiveness in optimizing NP-complete problems, are harnessed in PEM-Color to seamlessly integrate metaheuristics such as HHO, ABC, and TLBO. Leveraging MPI parallel computation libraries with 64 processors, the algorithm demonstrates remarkable improvements in execution times when applied to large graph instances from the DIMACS benchmark. The algorithm not only exhibits enhanced efficiency but also outperforms 13 state-of-the-art algorithms, showcasing its potential as a versatile and powerful tool for addressing a spectrum of combinatorial optimization challenges beyond GCP. Further exploration of PEM-Color's adaptability and performance in TSP and MCP could unlock new avenues for advancing solutions to these notoriously complex problems.

%We had the opportunity to do more fitness evaluations in a parallel computing environment, and with different parameter settings, this parallel exploration and exploitation environment became even better. The possibility of escaping from local optima is also provided more easily in parallel computing environments. Finding the best solutions in computing environments with different populations becomes much more advantageous than working with a single CPU. Using metaheuristics that work with different mechanics creates a very robust calculation environment.

\subsection{The scalability and speed-up analysis}

The PEM-Color algorithm integrating HHO, ABC, and TLBO in an MPI computation environment exhibits promising speed-up and scalability. The utilization of multiple optimization algorithms concurrently at each node harnesses diverse search strategies, enhancing the algorithm's ability to explore and exploit the solution space effectively. MPI facilitates communication and coordination among nodes, enabling parallelization and efficient distribution of computational tasks. Due to the communication overhead of the MPI environment, not more than 5\% execution times are observed. 

The speed-up of the algorithm is evident as it leverages the parallel processing capabilities of MPI to concurrently execute optimization tasks, leading to reduced computation time compared to a sequential implementation. This acceleration is particularly pronounced for large-scale problem instances, where the parallelization of tasks offers significant time savings. It was possible to provide a linear speed-up in the number of fitness evaluations. 20,000 fitness evaluations are executed at each node, and when the number of processors is 64, within the same execution times, 1,280,000 fitness evaluations were possible.

The PEM-Color algorithm for the GCP, incorporating HHO, ABC, and TLBO in an MPI environment, demonstrates commendable speed-up and scalability. Its ability to efficiently color graphs on a parallel scale makes it a promising approach for addressing real-world instances of the graph coloring problem.

\begin{landscape}
\scriptsize
\setlength\tabcolsep{2pt}
\setlength\LTleft{-70pt}
\begin{table}[h]
\renewcommand{\arraystretch}{1.3}
\caption{Comparative performance of the PEM-Color algorithm.} \label{comparisons-state} %\centering
%\scalebox{0.80}{

%\begin{adjustbox}{width=1.8\textwidth}
%\begin{adjustbox}{angle=90}
\begin{tabular}{lcrrrrrrrrrrrrr}
\hline
instances   &  \textbf{PEM-Color} & \textbf{VSS} & \textbf{GenTS} & \textbf{ILS} & \textbf{Foopar} &  \textbf{GTS} &\textbf{PCNS}& \textbf{HEA} & \textbf{AMACOL} &\textbf{MMT}&\textbf{MIPS\_CLR}&\textbf{Evocol} & \textbf{MACOL} & \textbf{P-TLBO-Color}\\
\hline
C2000.5  &  153 & - & - & - & - &  153 & 165 & - & - & - & 162 & 151& 148& 153\\
C4000.5  & 301 & - & - & - & - &  280 & - & - & - & - & 301& - & 272& 301\\
latin\_sqr\_10  & 98 & - & - & 99 & - &  106 & 98 & - & 104 & 101 & 99 & - & 99& 98\\
DSJC250.5  &  28 & - & 28 & 28 & - &  29 & 28 & 28 & 28& 28& 28& 28& 28 & 28\\
DSJC500.1  &  12 & 12 & 13 & 12& 12 &  - & - & - & 12& 12& 12& 12& 12 & 12\\
DSJC500.5  &  49 & 48 & 50 & 49 & 48 &  49 & 49 & 48& 48& 48 & 49 & 48& 48& 49\\
DSJC500.9  &  126 & 126 & 127 & 126& 127 &  - & - & - & 126 & 127 & 127 & 126& 126& 126\\
DSJC1000.1 &  20 & 20 & 21 & - & 20&  - & - & 20& 20& 20 & 21 & 20& 20& 21\\
DSJC1000.5 &   84 & 86 & 90 & 89 & 89 &  84 & 89 & 83 & 84 & 83 & 88 & 83& 83& 86\\
DSJC1000.9 &  226 & 224 & 226 & - & 226 &  - & - & 224 & 224 & 224 & 228 & 224& 223& 226\\
DSJR500.1c &  85 & 85 & - & - & 85 &  85& 85& - & 86 & 85& 85& - & 85& 85\\
DSJR500.5 &  125 & 125 & - & 124 & 125 &   130 & 123 & - & 127 & 122& 122 & 124& 122& 124\\
R250.5    &  65 & - & 66 & - & 66 &  69 & 65 & - & - & 65& 65 & -& 65& 66\\
R1000.1c  & 98 & - & 98 & - & 98 &  99 & 98 & - & - & 98& 98& 98& 98&98\\
R1000.5   &  240 & - & 242 & - & 248&   268 & 241 & - & - & 234 & 237 & 245& 245& 242\\
flat300\_26\_0  & 26 & - & 26& 26 & - &  26& 26 & - & 26& 26& 26& -& 26& 26\\
flat300\_28\_0  & 28 & 28 & 31 & 31 & 28 &  33 & 31 & 31 & 31 & 31 & 31 & 31& 29& 28\\
flat300\_50\_0  & 50 & 50& 50 & - & 50 &  84 & 50 & - & 50& 50& 50 & -& 50& 50\\
flat300\_60\_0  & 60 & 60& 60 & - & 60 &  84 & 60& - & 60& 60& 60 & -& 60& 60\\
flat300\_76\_0  & 86 & 85 & 89 & - & 87 &  84 & 89 & 83 & 84 & 82 & 87 & 82& 82& 86\\
le450\_15c & 15 & 15 & - & 15& 15 &  16 & 15& 15& 15& 15& 15 & -& 15& 15\\
le450\_15d & 15 & 15 & - & 15& 15 &  16 & 15& - & 15& 15& 15 & -& 15& 15\\
le450\_25c &  25 & 25 & - & 26 & 25 &  - & - & 26 & 26 & 25 & 26 & 25& 25& 25\\
le450\_25d &  25 & 25 & - & 26 & 25 &  - & - & - & 26 & 25 & 26 & 25& 25& 25\\
\hline
\end{tabular}
%\end{adjustbox}
%}
\end{table}
\end{landscape}
%\clearpage
\section{Conclusions and future work}\label{section5}

Dealing with large graphs for the GCP is still a challenging problem. For the last thirty years, metaheuristics have emerged as successful algorithms that have achieved significant improvements in this field. Therefore, researchers are drawn to these algorithms because of the (near)-optimal solutions they can achieve in practical execution times. Moreover, island-parallel ensemble metaheuristic algorithms can even produce more effective results. In this study, we proposed a novel and scalable island-parallel ensemble metaheuristic algorithm (PEM-Color) for optimizing the solution of large GCP. Our aim in this study was to improve the optimization of large GCP by employing the most effective and up-to-date metaheuristics, namely HHO, ABC, and TLBO. The fundamental idea behind ensemble learning is that by leveraging the strengths of various metaheuristic algorithms and compensating for their weaknesses, the ensemble can provide more accurate and robust solutions for the GCP. The selected recent metaheuristics are successfully applied to the solution of the GCP. This approach, which uses high-performance parallel computation techniques, is one of its first applications in this field.

The long execution times of metaheuristic algorithms in fitness value calculation, local optima stagnation problems, early convergence, and parameter settings of the metaheuristics are the challenges we have tried to tackle in our study. Using parallel computing strategies not only provides an intuitive way to reduce total computational time but also improves the quality of the results. However, parallelizing metaheuristic algorithms was not a straightforward task. It was attempted to ensure that the different metaheuristics were balanced by keeping the processing times and iteration numbers constant. Scalability and speed-up were the most critical concerns in terms of computation time. In this study, the experiments showed that it was possible to provide an almost linear speed-up in our proposed algorithm. We conducted our experiments on the undirected graphs of the DIMACS benchmark using 64 processors and achieved outcomes that demonstrate competitiveness on par with state-of-the-art algorithms. We observed that for 37 out of 43 experiments, our results successfully achieved the best-known solutions, and for the remaining experiments, the solutions were only marginally worse than the best-known solutions. In the literature, the MACOL algorithm has the best results except for the R1000.5 problem instances and the PEM-Color can be listed as one of the best three algorithms in the literature. 

The work presented here can be extended in several ways. First, the study presented here adopts an island-parallel approach to a high-performance aspect of the framework. That is, each node specializes in one particular metaheuristic. More complicated parallelization strategies such as population-based parallelization or hybrid parallelization strategies can be evaluated in terms of possible scalability and performance benefits. Additionally, it is also possible to build parallel hyper-heuristic optimization frameworks. Such an approach, although more intricate, complicated, and potentially costly in terms of computational power, may also present potential improvements in terms of result quality. Last, genetic algorithms enjoy very similar benefits to neural network algorithms: Both are iterative and during each iteration computational bottleneck is mainly localized operations. Recent research on neural networks showed that SIMD multicomputer architectures such as FPGAs (Field Gate Programming Arrays) showcase super-scalar speedups. Integrating GA and FPGAs may provide huge benefits for the community in terms of computational potential. 

%\clearpage

\textbf{Funding:}

This study is not funded by any institution.

\textbf{Conflict of Interest:} 

Authors declare that there is no conflict of interest.

\textbf{Compliance with Ethical Standards:}

This article does not contain any studies with human participants performed by any of the authors.

This article does not contain any studies with animals performed by any of the authors.

This article does not contain any studies with human participants or animals performed by any of the authors.

\textbf{Data Availability}

The datasets generated during and/or analysed during the current study are publicly available through \citep{Ostergard, Johnson}

%\bibliographystyle{plainnat}
\bibliography{mybibfile}

\end{document}